\documentclass[dvipsnames,format=sigconf,anonymous=false,review=false]{acmart}
\usepackage{graphicx} 
\usepackage{scalefnt}
\usepackage{adjustbox}
\usepackage{tikz}
\usepackage{forest}
\usetikzlibrary{shapes,arrows}
\usetikzlibrary{positioning}
\usepackage{algorithm}
\usepackage{algpseudocode}
\usepackage[color=lime]{todonotes}
\usepackage{listings}
\usepackage{enumitem}
\usepackage{hyperref}
\usepackage{subcaption}

\algrenewcommand\textproc{}

\makeatletter
\newcommand*{\algrule}[1][\algorithmicindent]{\makebox[#1][l]{\hspace*{.5em}\vrule height .75\baselineskip depth .25\baselineskip}}%

\newcount\ALG@printindent@tempcnta
\def\ALG@printindent{%
    \ifnum \theALG@nested>0
        \ifx\ALG@text\ALG@x@notext
            \addvspace{-3pt}
        \else
            \unskip
            \ALG@printindent@tempcnta=1
            \loop
                \algrule[\csname ALG@ind@\the\ALG@printindent@tempcnta\endcsname]%
                \advance \ALG@printindent@tempcnta 1
            \ifnum \ALG@printindent@tempcnta<\numexpr\theALG@nested+1\relax
            \repeat
        \fi
    \fi
    }%
    
\algnewcommand\algorithmicforeach{\textbf{for each}}
\algdef{S}[FOR]{ForEach}[1]{\algorithmicforeach\ #1\ \algorithmicdo}

\algblockdefx[Foreach]{Foreach}{EndForeach}[1]{\textbf{foreach} #1 \textbf{do}}{\textbf{end foreach}}

\setcopyright{acmcopyright}
\copyrightyear{2025}
\acmYear{2025}
\setcopyright{rightsretained}
\acmConference[GECCO '25 Companion]{Genetic and Evolutionary Computation
Conference}{July 14--18, 2025}{Malaga, Spain}
\acmBooktitle{Genetic and Evolutionary Computation Conference (GECCO '25
Companion), July 14--18, 2025, Malaga,
Spain}\acmDOI{10.1145/3712255.3726697}
\acmISBN{979-8-4007-1464-1/2025/07}
\acmPrice{15.00}

\author{Roman Kalkreuth}
\affiliation{%
    \institution{RWTH Aachen University}
    \city{Aachen}
    \country{Germany}}
\email{kalkreuth@aim.rwth-aachen.de}
\orcid{0000-0003-1449-5131}

\author{Fabricio Olivetti de França}
\affiliation{%
    \institution{Universida de Federal do ABC}
    \city{Santo André}
    \country{Brazil}}
\email{folivetti@ufabc.edu.br}
\orcid{0000-0002-2741-8736}

\author{Julian Dierkes}
\affiliation{%
    \institution{RWTH Aachen University}
    \city{Aachen}
    \country{Germany}}
\email{dierkes@aim.rwth-aachen.de}
\orcid{0009-0006-8670-398X}

\author{Marie Anastacio}
\affiliation{%
    \institution{RWTH Aachen University}
    \city{Aachen}
    \country{Germany}}
\email{anastacio@aim.rwth-aachen.de}
\orcid{0000-0002-4039-2470}

\author{Anja Jankovic}
\orcid{0000-0001-9267-4595}
\affiliation{%
    \institution{RWTH Aachen University}
    \city{Aachen}
    \country{Germany}}
\email{jankovic@aim.rwth-aachen.de}

\author{Zdenek Vasicek}
\affiliation{%
    \institution{Brno University of Technology}
    \city{Brno}
    \country{Czech Republic}}
\email{vasicek@fit.vutbr.cz}
\orcid{0000-0002-2279-5217}

\author{Holger Hoos}
\affiliation{%
    \institution{RWTH Aachen University}
    \city{Aachen}
    \country{Germany}}
\email{hh@aim.rwth-aachen.de}
\orcid{0000-0003-0629-0099}

\algnewcommand{\algorithmicand}{\textbf{ and }}
\algnewcommand{\algorithmicor}{\textbf{ or }}
\algnewcommand{\OR}{\algorithmicor}
\algnewcommand{\AND}{\algorithmicand}
\algnewcommand{\var}{\texttt}
\let\oldReturn\Return
\renewcommand{\Return}{\State\oldReturn}

\algnewcommand{\LineComment}[1]{\State \(\triangleright\) \textit{\color{blue} \scriptsize #1}}

\definecolor{hhcolor}{rgb}{0.2,0.6,0.6}
\definecolor{rkcolor}{rgb}{0, 0, 0.6} 
\definecolor{jdcolor}{rgb}{1.0, 0.6, 0.2}
\definecolor{ajcolor}{rgb}{0.7, 0.0, 0.3}
\definecolor{todocolor}{rgb}{0.9,0.1,0.1}
\definecolor{changedcolor}{rgb}{0.42,0.27,0.57}

\title{TinyverseGP: Towards a Modular Cross-domain Benchmarking Framework for Genetic Programming}
\author{}

\begin{abstract}
Over the years, genetic programming (GP) has evolved, with many proposed variations, especially in how they represent a solution. Being essentially a program synthesis algorithm, it is capable of tackling multiple problem domains. Current benchmarking initiatives are fragmented, as the different representations are not compared with each other and their performance is not measured across the different domains. In this work, we propose a unified framework, dubbed TinyverseGP (inspired by tinyGP), which provides support to multiple representations and problem domains, including symbolic regression, logic synthesis and policy search.
\end{abstract}

\begin{document}

\keywords{Genetic Programming, Implementation, Benchmarking, Symbolic Regression, Logic Synthesis, Python}

\maketitle
\renewcommand{\shortauthors}{Kalkreuth, Olivetti de Fran\c{c}a et al.}

\section{Introduction}

Genetic programming (GP) is an evolutionary algorithmic paradigm from the field of randomised search heuristics. 
It was first introduced by Cramer~\cite{cramer1985representation} for the automated discovery of programs, and later popularised by Koza, with his GP-driven search to hierarchical structures~\cite{Koza:1992:GPP:138936}, which led to meaningful applications in symbolic regression, one of the major problem domain of GP. 
Other problem domains have emerged, notably classification and digital circuit design, and revealed the need for different solution encodings, such as linear~\cite{ieee94:perkis}, graph-based~\cite{DBLP:conf/eurogp/MillerT00}
and grammar-guided~\cite{ryan:1998:geepal} representations.
The interplay between new representations and applications has led to a versatile landscape, but also to fragmentation and encapsulation, hindering the development of unified and cross-domain knowledge. 
Obsolete real-world problems have been proposed as benchmarks and bundled in domain-specific benchmark suites for program synthesis, symbolic regression and logic synthesis. 
However, cross-domain evaluation entails a large implementation overhead.

In this work, we take a step towards a unified framework, combining representation models and benchmarks. Our approach, dubbed \textit{TinyverseGP}, is inspired by tinyGP~\cite{Sipper2019tinyGP}, following the philosophy of keeping a minimalist implementation of each variant. 
The shared concepts are kept in a common library, making it easy to integrate new representations and to support different problem domains.
We intend to create a collaborative space to incorporate the different variants and problem domains, facilitating the understanding of the characteristics of each representation. 
As a starting point, we showcase the integration of a tree-based and graph-based GP in the domains of logic synthesis, symbolic regression, and policy search.

\section{Related Work}
\label{sec:related_work}
\paragraph{\bf Genetic Programming}
GP is an evolutionary search methodology, located in the wider field of randomised search heuristics, and was originally proposed for the synthesis of computer programs. 
GP as a heuristic paradigm aims at \textit{evolving} a population of candidate programs towards a functionally specified
solution for a given search problem. 
Fundamental to GP-driven search is the iterative transformation of a population of candidates into new populations consisting of programs with an improved fitness score. 
To represent a program, GP traditionally uses parse trees, which have been inspired by LISP S-expressions. 
Commonly used variation operators applied to the traditional tree-based representation are subtree crossover and mutation. Subtree crossover swaps subtrees between two trees selected for recombination, while subtree mutation exchanges a subtree with a randomly generated one. 

\paragraph{\bf Problem Domains in GP}
Because the very essence of GP is to search for a computer program, it can be readily applied to different machine learning tasks that would normally require major adaptations. 
In most cases, merely a grammatical specification for the respective task is required to seek a solution with GP paradigm.
For example, in regression and classification tasks, the grammar can be composed of mathematical operators and conditional branching. 
Likewise, in policy search, GP can search for the function that estimates the value of a given action at the current state, reverting to a symbolic regression problem, or a computer program that will return the optimal action for that particular state, thus incorporating branching and loops in the grammar. 
Additionally, GP can also search for Boolean functions, if we constrain the grammar to the corresponding set of operators, a problem that is often performed with heuristics or exhaustive search.

\paragraph{\it Symbolic Regression}
Symbolic regression (SR)~\cite{Koza:1992:GPP:138936} searches for a mathematical expression representing a regression model that accurately and compactly describes the data. Given a set of $m$ points $\{x_i, y_i\}_{i=1 \ldots m}$, this task seeks the function $f(x) \approx y$. 

\paragraph{\it Logic Synthesis}
Logic synthesis (LS)~\cite{DBLP:books/daglib/0086041}, as approached with heuristic methods, can be defined as a black-box optimisation task that aims at the synthesis of Boolean expressions.
As in symbolic regression, the expression to be synthesised has to match the correct input-output mapping of a (Boolean) function. 
LS tackled with GP predominantly aims at synthesis of Boolean expressions that match the correct input-output mapping of a Boolean functions. 

\paragraph{\it Policy Search}
Policy search focuses on finding decision-making strategies that maximise performance through interaction with an environment \cite{10.5555/3312046}. 
The environment is formalised as a Markov decision process (MDP) $\mathcal{M} := (\mathcal{S}, \mathcal{A}, p, r, \rho_0, \gamma)$, with state space $\mathcal{S}$, action space $\mathcal{A}$, unknown transition probability distribution $\rho: \mathcal{S} \times \mathcal{A} \times \mathcal{S} \mapsto \mathbb{R}$, reward function $r: \mathcal{S} \times \mathcal{A} \mapsto \mathbb{R}$, distribution of the initial state $\rho_0: \mathcal{S} \mapsto \mathbb{R}$ and discount rate $\gamma \in (0, 1)$.
A policy \( \pi: \mathcal{S} \times \mathcal{A} \to \mathbb{R} \) assigns a probability to each action for a given state. 
Interacting with the MDP, $\pi$ collects episodes $\tau = (s_0, a_0, r_1, s_1, \cdots, s_T)$ over time steps $t = 0,\cdots, T$.
The fitness of a policy is measured as the return $\mathbb{E}_{\tau \sim \pi}[\sum^T_{t=1}\gamma^t \cdot r_t]$.

\section{Benchmarking in Genetic Programming}
GP benchmarks have, for many years, been criticised by their lack of rigour, caused by the fragmentation of the field~\cite{McDermott:2012:GECCO}. 
It has been argued that a first step towards better benchmarks is the standardisation to ensure a challenging environment and fair comparison between different flavours of GP as well as alternatives from machine learning. 
A major step forward in GP benchmarking has been made by the proposal of two benchmark suites for program synthesis, PSB1~\cite{DBLP:conf/gecco/HelmuthS15} and PSB2~\cite{DBLP:conf/gecco/HelmuthK21}, which cover a diverse set of introductory and college-level coding problems. 
Moreover, benchmark suites for other problem domains, such as SRBench~\cite{Orzechowski:2018:GECCO} and GBFS~\cite{DBLP:conf/foga/KalkreuthVHVYB23}, have emerged, and new problem domains have been discovered in recent years~\cite{10.1145/3520304.3528766}. 

\paragraph{\bf SRBench: A Living Benchmark for Symbolic Regression}

Orzechowski et al.~\cite{Orzechowski:2018:GECCO} compared a selection of Symbolic Regression (SR) with traditional machine learning techniques, addressing a common criticism regarding the lack of comparison between these approaches.
Their results indicated that SR was a competitive alternative to opaque models in terms accuracy, with the benefit of returning interpretable models. La Cava et al.~\cite{LaCava:2021:NeurIPS} advanced this effort even further by incorporating new algorithms into the benchmark, dubbed SRBench, and proposing a collaborative environment facilitating the benchmarking of new algorithms using a common Python interface and a verified installation environment that enabled external peers to replicate the benchmark. 
This effort spanned multiple competitions, one of which led to a publication highlighting the challenges still faced by the field of SR in general, not only GP~\cite{deFranca:ieeeTEC}. 

\paragraph{\bf GBFS: General Boolean Function Benchmark Suite}

The General Boolean Function Benchmark Suite (GBFS)~\cite{DBLP:conf/foga/KalkreuthVHVYB23} is a highly versatile benchmark suite for logic synthesis proposed with the intention to facilitate comprehensive assessment of the performance of GP models as well as to simplify reproducibility of existing results in this problem domain.
GBFS covers a set of $29$ problems carefully selected from seven different types of Boolean functions: arithmetic, transmission, comparison, counting, mixed, parity and cryptography. 
Most of the problems are characterised by having multiple outputs, such as the digital adder or multiplier function.

\paragraph{\bf Reinforcement and Policy Learning}
A variety of benchmarks have been developed and proposed for Reinforcement Learning (RL) in recent years, tailored to diverse use cases such as games, planning and robotics \cite{bellemare13arcade, MinigridMiniworld23, robosuite2020}. 
A widely used platform is Gymnasium \cite{towers2024gymnasium}, which provides diverse environments aimed at evaluating the core capabilities of RL algorithms.
Additionally, the Atari Learning Environment (ALE) \cite{bellemare13arcade} offers a standardised and challenging testbed for RL agents.
Benchmarking GP on Gymnasium has already been explored in \cite{10.1145/3520304.3528766}.

\section{The Proposed Framework}

\paragraph{\bf Major Motivation}
The landscape of existing GP frameworks can be considered fragmented, 
in terms of representation as well as application domain. 
We acknowledge the implementation efforts made over time to facilitate broad use of GP. 
However, existing implementations are meant to provide an end-user experience, thus reducing the burden of choosing the most suitable representation for each task.
Notably, there is clearly a gap on benchmarking and understanding the different representations proposed so far across different domains, as evidenced by the paper \emph{Genetic programming needs better benchmarks}~\cite{McDermott:2012:GECCO}, which received the 2022 SIGEVO Impact Award and led to initiatives such as SRBench~\cite{LaCava:2021:NeurIPS} and GBFS~\cite{DBLP:conf/foga/KalkreuthVHVYB23}, which provide tight guidelines to ensure fairness of comparison among the competing algorithms.

\paragraph{\bf Key Features and Properties}
For the first version, we concentrated on the development of the following features and properties that represent the fundamental infrastructure of TinyverseGP. 
In general, we pursue object-oriented design, to ensure that the design goals communicated in the previous subsection can be adhered to. 

\begin{itemize}[leftmargin=*]
\item \textbf{Light-weight representation modules:} Each representation derives from a model base class, which promotes uniformity among different representations. Each representation is implemented within a tiny module that handles initialisation, decoding, breeding and evaluation of candidate programs concerning the requirements of the respective model. 
\item 
\textbf{Broad range of applications:} TinyverseGP can already be  applied to several problems that differ greatly and require different methods on how to evaluate an evolved program, showing the versatility of this tool.
\item \textbf{Benchmarking support:} We provide an interface to SRBench and GBFS, covering two problem domains in combinational synthesis. 
Additionally, we provide an interface to policy learning benchmarks from Gymnasium, which is accomplished with an implementation of an agent class that bridges the gap between the respective GP model and the environment. 
\end{itemize}

\section{Top-level Architecture}
TinyverseGP hybridises object-oriented design with modularisation, two programming paradigms commonly used to establish an efficient reusable architecture.
With modularisation, we also facilitate encapsulation of diverse techniques and methodologies, to maintain an organised and interpretable code structure, as TinyverseGP is expected to become larger and more heterogeneous in the future. 
We therefore leverage modularisation and object-oriented features to implement a design that is well-equipped to apply various GP models to a diverse set of benchmarks.

\begin{figure} 
  \includegraphics[scale=0.5]{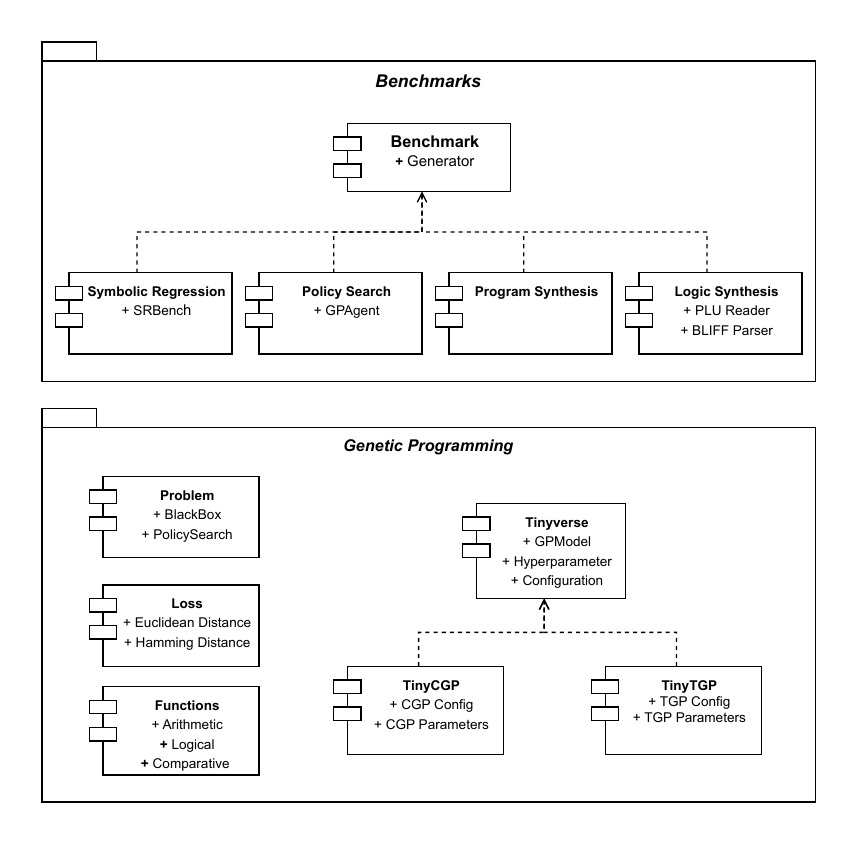}
  \caption{Modular top-level view of TinyverseGP}
  \label{img:modular_architecture}
\end{figure}

Figure~\ref{img:modular_architecture} provides an overview of the modular architecture of TinyverseGP. 
The GP module represents the core of our framework, and the current architecture allows further integration with less effort; existing Python implementations of other GP models can be easily integrated into the already existing framework model, which already provides essential features to handle function and terminal sets, as well as the hyperparameter and model configuration. 

\begin{figure*}
\centering
\begin{minipage}[b]{.6\textwidth}
  \includegraphics[scale=0.51]{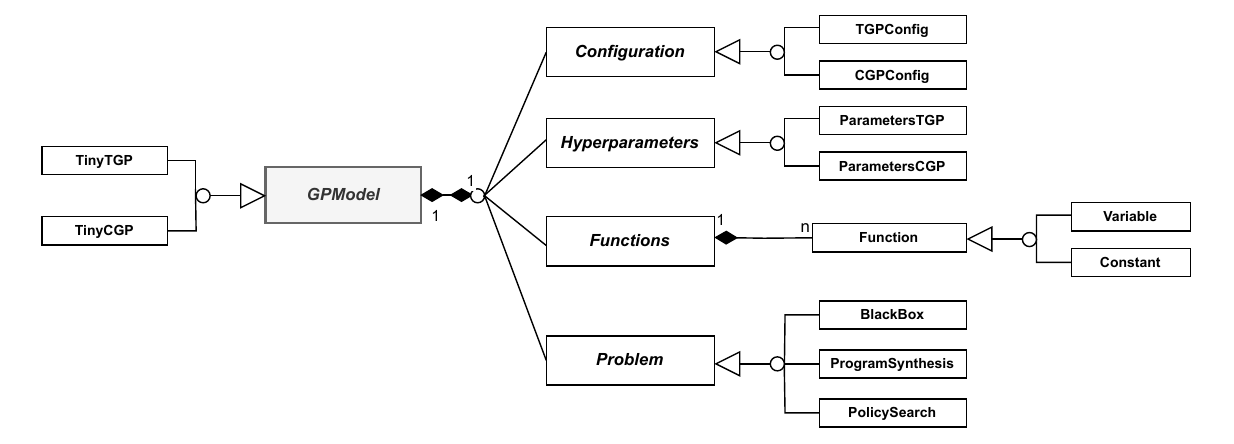}
\caption{High-level architecture of TinyverseGP}\label{label-a}
\label{img:class_architecture}
\end{minipage}
\begin{minipage}[b]{.37\textwidth}
 \includegraphics[scale=0.44]{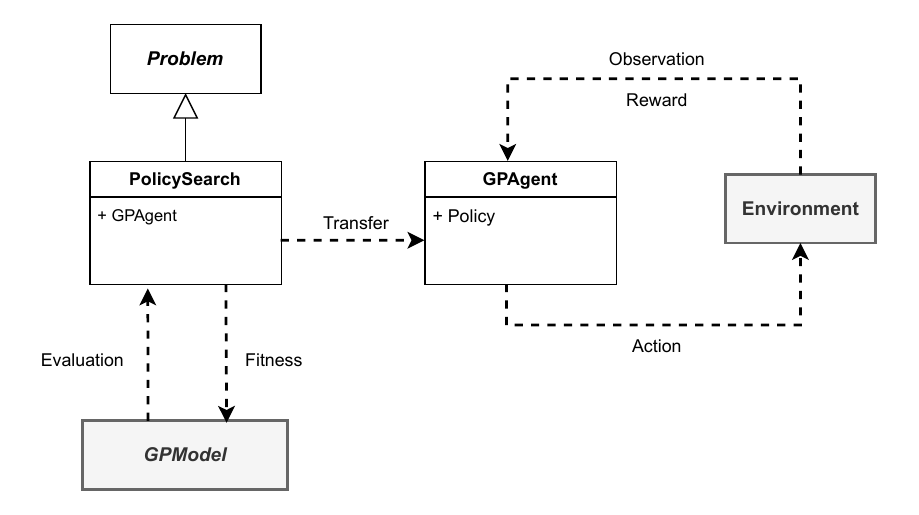}
\caption{Policy search with TinyverseGP}\label{label-b}
\label{img:policy_search}
\end{minipage}
\end{figure*}

The fundamental architecture is illustrated in Figure~\ref{img:class_architecture} on the class level. 
The most common problem type that is tackled with the GP paradigm are black-box problems, where known input-output pairs are given in the respective training dataset. 
Policy search problems, on the other hand, are commonly applied to environments where an agent executes actions in accordance to a policy for which he receives rewards. For this type of problem we provide an agent class that receives a candidate policy from the respective GP model, as illustrated in Figure~\ref{img:policy_search}.

\section{Discussion, Vision and Future Work}

The primary intention behind TinyverseGP is to take the first step towards a community-driven framework that simplifies the process of performing comparative studies among various representations of GP,
thereby promoting best practices for benchmarking randomised search heuristics within this field. 
We acknowledge the efforts that have already been made in this direction by the proposal of various benchmark suites for major GP problem domains.
We believe that facilitating cross-domain benchmarking that leverages previous efforts for specific domains is a natural next step and will bring the field forward.
However, unifying both the representation models and a diverse landscape of application domains in one framework is a task that requires well-considered design choices to be equipped for further expansion. 
Our vision for TinyverseGP can be summarised as follows:

\begin{itemize}[leftmargin=*]
    \item \textbf{Community driven development:} The source code will be open-sourced and freely available. 
    \item \textbf{Easy to extend:} By providing a common API requiring just a few methods, the framework can be easily extended without the need for understanding the entire codebase. 
    \item \textbf{Simplicity of implementation:} The minimalistic approach allows us a direct comparison between representations without the influence of other external agents that can influence the performance of the algorithm, such as different selection mechanisms, island models and local search. 
\end{itemize}

\def\checkmark{\tikz\fill[scale=0.2](0,.35) -- (.25,0) -- (1,.7) -- (.25,.15) -- cycle;} 

\begin{table}[ht!]
    \centering
    \caption{Planned support for different benchmarks and problem domains.}\label{tab:problems}
    \scalebox{0.8}{
    \begin{tabular}{ccp{3.5cm}}
    \toprule 
        \textbf{Problem Domains} & \textbf{Support} & \textbf{Benchmarks}  \\
        \midrule 
        Logic Synthesis & \checkmark & Classic, GBFS \\
        Symbolic Regression & (\checkmark) & Classic, SRBench, Feynman \\
        Policy Search & (\checkmark) & Gymnasium \\
        Program Synthesis & $\times$ & Leet Code, PBS 1 \& 2, SyGuS  \\ 
        \bottomrule 
    \end{tabular}
    }
\label{tabl:supported}
\end{table}

\paragraph{\bf Planned Features and Extensions}
Two specific extensions can be considered as natural next steps for the project. 
Firstly, we plan to include linear-based and grammar-based GP in the context of our tiny module approach, enabling a scope on the representation level that can be used for a first broad comparative study performed with TinyverseGP. 
Secondly, on the domain level, we will concentrate on the integration of program synthesis benchmarks by providing an interface for the General Program Synthesis Benchmark Suites (PSB1 and PSB2) proposed by Helmuth \emph{et al.}~\cite{DBLP:conf/gecco/HelmuthS15, DBLP:conf/gecco/HelmuthK21}; 
a summary is given in Table~\ref{tabl:supported}, where
supported benchmarks are marked using \checkmark, partially supported ones using (\checkmark), and those under development using $\times$.

\section{Conclusion}
\label{sec:conclusions}

We have proposed the first prototype of a modular cross-domain benchmarking framework for GP that is meant to be the blueprint for a large and scalable framework.
TinyverseGP can already be used to evaluate various benchmarks from three problem domains to which GP is highly applicable, including the underrepresented domain of policy search. 
to achieve a better balance against overused benchmarks.

\section{Resources}
\label{sec:resources}
The source code of the actively developed main branch, test scripts and a handbook are available on GitHub.\footnote{\href{https://github.com/GPBench/TinyverseGP}{https://github.com/GPBench/TinyverseGP}} 
The version proposed in this paper is provided in a locked branch.\footnote{\href{https://github.com/GPBench/TinyverseGP/tree/gecco-2025}{https://github.com/GPBench/TinyverseGP/tree/gecco-2025}}

\section{Acknowledgements}
\label{sec:acknowledgement}
This work was supported by an Alexander von Humboldt Professorship in AI held by Holger Hoos, Czech Science Foundation project 25-15490S and Conselho Nacional de Desenvolvimento Cient\'{i}fico e Tecnol\'{o}gico (CNPq) grant 301596/2022-0.


\bibliographystyle{ACM-Reference-Format}
\bibliography{bib/gecco25_poster}

\end{document}